\documentclass{article}

\usepackage{PRIMEarxiv}

\usepackage[utf8]{inputenc} 
\usepackage[T1]{fontenc}    
\usepackage[authoryear,round]{natbib} 
\usepackage{hyperref}       
\usepackage{url}            
\usepackage{booktabs}       
\usepackage{amsfonts}       
\usepackage{nicefrac}       
\usepackage{microtype}      
\usepackage{lipsum}
\usepackage{fancyhdr}       
\usepackage{graphicx}       
\usepackage{multirow}   
\usepackage{makecell}   
\usepackage[table]{xcolor} 
\usepackage{colortbl} 
\usepackage{subcaption} 
\usepackage{caption}
\usepackage{enumitem}

\usepackage[ruled,vlined]{algorithm2e}
\usepackage{algorithmic}
\usepackage{amsmath, amssymb}
\graphicspath{{media/}}     

\SetKw{KwBy}{by}

\title{A Case Study of Selected PTQ Baselines for Reasoning LLMs on Ascend NPU}
\author{
  Yuchen Luo$^{1}$, Fangyue Zhu$^{1}$, Ruining Zhou$^{1}$, Mingzhe Huang$^{1}$, Jian Zhu$^{1}$, Fanyu Fan$^{1}$, Wei Shao$^{2}$ \thanks{corresponding author}\\
  {\normalsize $^{1}$School of Mathematics and Statistics, Wuhan University, Wuhan, China} \\
  {\normalsize $^{2}$Huawei} \\
  {\normalsize \{\texttt{yuchenluo,zhufangyue,2020302011111,2025202010028,jianzhu,fyfang}\}@whu.edu.cn} \\
  {shaowei99@huawei.com} \\
}

\begin{document}
\maketitle

\begin{abstract}
Post-Training Quantization (PTQ) is crucial for efficient model deployment, yet its effectiveness on Huawei’s Ascend NPUs remains under-explored compared to GPU architectures. This paper presents a case study of representative PTQ baselines applied to reasoning-oriented models such as DeepSeek-R1-Distill-Qwen series (1.5B/7B/14B) and QwQ-32B. We evaluate four distinct algorithms, including AWQ, GPTQ, SmoothQuant, and FlatQuant, to cover the spectrum from weight-only compression to advanced rotation-based methods. Our empirical results reveal significant platform sensitivity. While 4-bit weight-only quantization proves viable for larger models, aggressive 4-bit weight-activation schemes suffer from layer-wise calibration instability on the NPU, leading to logic collapse in long-context reasoning tasks. Conversely, standard 8-bit quantization remains numerically stable. Furthermore, a real-world INT8 deployment demonstrates that although optimized kernels reduce latency, dynamic quantization overheads currently limit end-to-end acceleration. These findings offer a practical reference for the feasibility and limitations of deploying quantized reasoning models on Ascend NPU.
\end{abstract}


\section{Introduction}
Large language models (LLMs) \citep{grattafiori2024llama,yang2025qwen3,guo2025deepseek} have revolutionized natural language tasks, yet their massive parameter counts impose prohibitive constraints on memory and deployment. Quantization serves as a pivotal solution \citep{lin2024awq,frantar2022gptq}, though it risks degrading the model's reasoning capabilities \citep{liu2025quantization,li2025quantization}. Crucially, current quantization research is primarily focused on NVIDIA GPU architectures, leaving a gap in understanding efficacy on alternative hardware like Huawei’s Ascend NPU \citep{zuo2025serving}.

Deploying quantized LLMs on Ascend presents distinct hardware-software challenges. While NVIDIA GPUs benefit from mature libraries (e.g., BitBLAS \citep{ladder-osdi24}, Marlin \citep{frantar2025marlin}), Ascend lacks native support for specific operators required by state-of-the-art algorithms. For instance, advanced rotation-based methods like QTIP \citep{tseng2024qtip}, Quip\# \citep{tseng2024quip}, QuaRot~\citep{ashkboos2024quarot}, SpinQuant~\citep{liu2024spinquant}, and Ostquant~\citep{hu2025ostquant} depend on the Fast Hadamard Transform (FHT) \citep{pytorch2024hadacore,dao2024fht}. The absence of optimized FHT implementations on Ascend currently limits the practicality of these methods, motivating a re-evaluation of quantization strategies tailored to the NPU's constraints.

Given the Ascend architecture's distinct compute primitives (e.g., specific constraints on mixed-precision accumulation) and the lack of native support for certain outlier-suppression operators, a direct transfer of GPU-based conclusions is insufficient. To address this, we conduct a systematic evaluation of PTQ methods on Ascend, carefully selecting algorithms that align with its architectural constraints. We adopt a two-tiered selection strategy:

First, we establish AWQ~\citep{lin2024awq}, GPTQ~\citep{frantar2022gptq}, and SmoothQuant~\citep{xiao2023smoothquant} as our primary baselines. Among them, AWQ and GPTQ represent the weight-only setting.
SmoothQuant, on the other hand, is originally a W8A8 method (with the KV cache typically kept in FP16), and in our experiments we additionally quantize the KV cache to 8-bit (denoted as W8A8KV8).
This selection is aligned with practical deployment on Ascend, where real INT8 quantized inference (e.g., W8A8) has already been demonstrated \citep{zuo2025serving}, and recent work further provides an efficient W4A16 implementation \citep{he2026w4a16}. Together, these results suggest that both W8A8 and weight-only low-bit inference on Ascend can be supported by optimized kernels, making them reasonable and reproducible baselines.

Second, distinct from FHT-based approaches, we incorporate FlatQuant~\citep{sun2024flatquant} as a representative of high-accuracy, rotation-based quantization. Crucially, FlatQuant utilizes standard kernel abstractions (e.g., OpenAI Triton \citep{tillet2019triton} and CUTLASS \citep{cutlass-gemm}) rather than specialized transforms. This design renders it potentially adaptable to the Ascend ecosystem via emerging toolchains like Ascend-Triton and CATLASS \citep{ascend2026tritonascend,huawei2026catlass}. In this study, we employ FlatQuant as a strategic proxy to estimate an approximate algorithmic upper bound of rotation-based quantization on NPU architectures. Our evaluation characterizes the accuracy of Triton-compatible quantization methods in isolation from kernel fusion complexities. 
We thereby separate algorithmic effects from kernel-engineering factors and help prioritize subsequent kernel-migration efforts.

In this study, we evaluate these quantization methods on both reasoning-focused and QA-focused settings.
For reasoning, we consider the open-source DeepSeek-R1-Distill-Qwen models \citep{guo2025deepseek} and QwQ-32B \citep{qwen2026qwq32b}, and benchmark them on AIME-120 \citep{maxwelljia2024aime}, MATH-500 \citep{lightman2023let}, GSM8K \citep{cobbe2021training}, GPQA-Diamond \citep{rein2024gpqa}, and LiveCodeBench \citep{jain2024livecodebench}.
For QA, we evaluate Llama-3 \citep{grattafiori2024llama} on ARC-Easy/ARC-Challenge \citep{clark2018think}, HellaSwag \citep{zellers2019hellaswag}, LAMBADA \citep{paperno2016lambada}, PIQA \citep{bisk2020piqa}, and WinoGrande \citep{sakaguchi2021winogrande}.

We summarize our main takeaways as follows:

    1. Under fake-quantized inference, 4-bit weight-only quantization (AWQ/GPTQ, w4g128) can be near-lossless or acceptable, but typically only for sufficiently large models (\(\geq\) 7B) on Ascend NPUs, while 3-bit weight-only settings (w3g128) often trigger severe accuracy collapse, especially at smaller scales \citep{lin2024awq,frantar2022gptq,liu2025quantization}.

    2. At 8-bit precision, SmoothQuant (W8A8KV8) and FlatQuant-W8A8KV8 remain numerically stable on Ascend and preserve downstream performance across both QA and reasoning benchmarks, making 8-bit a reliable operating point for practical NPU deployment \citep{xiao2023smoothquant,sun2024flatquant}.

    3. FlatQuant-W4A4KV4 exhibits strong platform sensitivity on Ascend: on QA benchmarks, performance can be largely recovered with targeted hyperparameter tuning, whereas long-context reasoning on distilled models still suffers persistent logic collapse (with QwQ-32B being relatively more robust) \citep{sun2024flatquant,liu2025quantization}.

    4. Beyond simulation, we implement a partial real-quantized INT8 inference path on Ascend to validate feasibility and identify bottlenecks. While integrating INT8$\times$INT8 matmul kernels can reduce latency for certain components, the current prototype does not yet fully leverage all available operators and is still constrained by dynamic quantization overheads, pointing to clear optimization opportunities.

Motivated by recent studies on quantization for reasoning models (e.g., \citet{liu2025quantization}), we conduct an empirical evaluation on Ascend NPUs by adapting representative PTQ baselines and carefully reproducing key experimental settings. We hope the resulting observations and comparisons provide a practical reference for quantizing and deploying reasoning-oriented models on NPUs.

\section{Quantization Frameworks and Empirical Paradigms}
Quantization is the process of compressing high-precision floating-point values in a tensor $\mathbf{X} \in \mathbb{R}^{m \times n}$ into lower-precision representations to reduce memory footprints and computational costs while striving to preserve model fidelity.  A standard formulation for uniform quantization represents the quantized approximation $\hat{\mathbf{X}}$ as:
\begin{align*}
\hat{\mathbf{X}} = \mathcal{Q}(\mathbf{X}; b) = \Delta \cdot \Pi_{\Omega(b)}\left(\frac{\mathbf{X}}{\Delta}\right),
\end{align*}

where $\Delta$ denotes the quantization step size (or scaling factor), and $\Pi_{\Omega(b)}(\cdot)$ is a projection function mapping values to the closest integer in the codebook $\Omega(b)$ defined by the bit-width $b$. 


A primary challenge in quantizing pre-trained LLMs is the presence of \textit{outlier channels} that exhibit magnitudes significantly larger than the surrounding distribution \citep{lin2024duquant,dettmers2022gpt3}. These outliers dominate the dynamic range, causing severe precision loss for non-outlier elements. Prior research broadly classifies solutions into three categories: storing outliers separately in high precision~\citep{dettmers2022gpt3,kim2024squeezellm,zhao2024atom}, designing non-uniform quantization algorithms \citep{shao2024omniquant,zhao2024atom,tseng2024qtip}, and applying equivalent weight transforms \citep{ashkboos2024quarot,sun2024flatquant,van2025fptquant,liang2025paroquant,he2025baseQ}.




Among the various PTQ methods currently available, transforming the distribution of weights or activations before quantization has proven to be an effective approach. For a linear layer \(\mathbf{Y} = \mathbf{X}\mathbf{W} + \mathbf{b}\), where input \(\mathbf{X} \in \mathbb{R}^{T \times D_{\mathrm{in}}}\), weight \(\mathbf{W} \in \mathbb{R}^{D_{\mathrm{in}} \times D_{\mathrm{out}}}\), and bias \(\mathbf{b} \in \mathbb{R}^{1 \times D_{\mathrm{out}}}\), we can apply an invertible transform \(\mathbf{T}\) to the weight \(\mathbf{W}\) without affecting the output:
\begin{equation*}
\mathbf{Y} = \mathbf{X}\mathbf{W} + \mathbf{b} = (\mathbf{X}\mathbf{T}^{-1})(\mathbf{T}\mathbf{W}) + \mathbf{b}.
\end{equation*}
For activations, we quantize \(\mathbf{X}\mathbf{T}^{-1}\), and for weights, we quantize \(\mathbf{T}\mathbf{W}\). A suitable \(\mathbf{T}\) can improve the distribution of both weights and activations, making them more uniform and reducing the presence of outliers, thereby reducing quantization error.

Previous research has mainly focused on two categories of transforms: channel-wise scaling (when \(\mathbf{T}\) is diagonal) and rotation (when \(\mathbf{T}\) is orthogonal). Channel-wise scaling scales each channel separately to even out the magnitude across channels and can usually be merged into preceding operators without incurring extra overhead \citep{lin2024awq,xiao2023smoothquant,touvron2023llama,wei2022outlier}. Rotation enables cross-channel interactions that can concentrate values more effectively than channel-wise scaling \citep{sun2024flatquant,akhondzadeh2025kurtail,yuan2023rptq,lin2024duquant,tseng2024quip,shao2025dartquant}. However, rotations cannot be merged into element-wise operators (e.g., layer normalization) like channel-wise scaling does, so they usually require online computation. In addition, there are also efficient quantization algorithms such as GPTQ~\citep{frantar2022gptq} and GPTAQ~\citep{li2025gptaq}, which utilize approximate second-order Hessian information to perform one-shot weight-only quantization and error compensation.
 

\section{Evaluation of Quantized Reasoning Models on ATLAS A2}
\label{sec:experimental_part}
\subsection{Experimental Setup}
\textbf{Models and Benchmarks}. Experiments are conducted on two categories of tasks using established evaluation protocols.
For question answering (QA), we evaluate on ARC-Easy and ARC-Challenge \citep{clark2018think}, HellaSwag \citep{zellers2019hellaswag}, LAMBADA \citep{paperno2016lambada}, PIQA \citep{bisk2020piqa}, and WinoGrande \citep{sakaguchi2021winogrande}. Baseline comparisons on LLaMA-3-8B follow the setup of \citet{sun2024flatquant}.
For mathematical and logical reasoning, we evaluate quantized models from the DeepSeek-R1-Distill-Qwen series \citep{hui2024qwen2} (1.5B, 7B, and 14B) as well as QwQ-32B \citep{qwen2026qwq32b}. The evaluation suite spans three domains: mathematical reasoning (GSM8K \citep{cobbe2021training}, MATH-500 \citep{lightman2023let}, and AIME-120 \citep{maxwelljia2024aime}), code generation (LiveCodeBench \citep{jain2024livecodebench}), and expert-level scientific reasoning (GPQA-Diamond \citep{rein2024gpqa}).
Evaluation for question answering tasks is conducted using Lighteval \citep{lighteval}, following the default generation and evaluation settings provided by the FlatQuant codebase to ensure consistency with prior work.
For mathematical and logical reasoning tasks, we employ vLLM-Ascend \citep{vllm2026vllmascend} to support efficient inference and long-context generation required by reasoning benchmarks. For reasoning evaluations, we align with the QHS baseline configuration in \citet{liu2025quantization}, adopting a sampling temperature of 0.6 and top-p of 0.95 during generation. The maximum sequence length is set to 32,768 tokens to accommodate the extended reasoning chains characteristic of these benchmarks.

\textbf{Quantization Strategies}. We benchmark a spectrum of quantization configurations ranging from weight-only compression to full quantization. For weight-only quantization, we employ AWQ and GPTQ with group size 128, evaluating both 3-bit and 4-bit configurations (W3A16g128, W4A16g128) to test the limits of parameter compression. For weight-activation quantization, we establish SmoothQuant (W8A8KV8) as the primary baseline. Additionally, we incorporate FlatQuant to evaluate the impact of advanced rotation-based quantization, testing both W8A8KV8 and the ultra-low-bit W4A4KV4 settings. This selection allows us to analyze the performance-efficiency trade-offs across different precision levels and quantization paradigms on the Ascend architecture.

Model quality is assessed using perplexity (PPL) together with task-specific accuracy or exact-match metrics.

\textbf{Device.} Our experiments are conducted primarily on the Ascend 910B NPU platform. Furthermore, for comparative purposes, we provide a baseline by including results from a device with an FP16 peak performance of approximately 2000 TFLOPS, which is referred to as X2000 in the subsequent sections.


\subsection{Key Observations}
\label{observation}

The comprehensive results for the DeepSeek-R1-Distill-Qwen series and QwQ-32B on the Ascend NPU are summarized in Table \ref{tab:reason}. Following the categorization framework established in \citep{liu2025quantization}, we classify performance degradation into three distinct levels: lossless (\textcolor[HTML]{F0C869}{orange} cells, $\leq 1\%$), fair (\textcolor[HTML]{A3A19D}{gray} cells, $1\%-3\%$), and risky (\textcolor[HTML]{9C9DDB}{blue} cells, $\geq 3\%$).

Our experimental results on the Ascend NPU partially echo the findings reported in \citep{liu2025quantization}, while revealing unique platform-specific challenges.
\begin{table*}[!t]
\centering 
\resizebox{1.00\linewidth}{!}{
    \begin{tabular}{c|l|l|rrrrr|r|r}  
    \hline\hline
    \textbf{Model} 
    & \textbf{Methods} 
    & \textbf{\makecell{W-A-KV\\\# Bits}} 
    & \textbf{\makecell{AIME-\\120}}  
    & \textbf{\makecell{MATH-\\500}} 
    & \textbf{GSM8K} 
    & \textbf{\makecell{GPQA-\\Diamond}} 
    & \textbf{\makecell{LiveCode-\\Bench}} 
    & \textbf{Avg.} 
    & \textbf{Drop}$\downarrow$ \\ 
    \hline
    \multirow{8}{*}{\rotatebox[origin=c]{90}{\textbf{\makecell{DeepSeek-R1\\-Distill-Qwen1.5B}}}}
    & \textbf{BF16} 
    & - 
    & \textbf{26.67} 
    & \textbf{86.20} 
    & \textbf{85.60} 
    & \textbf{37.88} 
    & \textbf{18.28} 
    & \textbf{50.92} 
    & \textbf{-} \\
    \cline{2-10}
    & \multirow{2}{*}{\textbf{\textit{awq}}} 
    & \multirow{1}{*}{4-16-16} 
    & 20.83
    & 81.00
    & 82.94
    & 32.32
    & 15.3
    & 46.48
    & \cellcolor{blue!12}{-4.44} \\  
    \cline{3-10}
    &  & \multirow{1}{*}{3-16-16} 
    & 5.83
    & 43.40
    & 62.47
    & 26.77
    & 2.61
    & 28.22
    & \cellcolor{blue!12}{-22.70} \\  
    \cline{2-10}
    
    & \multirow{1}{*}{\textbf{\textit{smoothquant}}} 
    & \multirow{1}{*}{8-8-8} 
    & 13.33
    & 75.40
    & 82.11
    & 32.32
    & 13.43
    & 43.32
    & \cellcolor{blue!12}{-7.60} \\ 
    \cline{2-10}
    
    & \multirow{2}{*}{\textbf{\textit{gptq}}} 
    & \multirow{1}{*}{4-16-16} 
    & 19.17
    & 83.60
    & 83.78
    & 34.85
    & 14.55
    & 47.19
    & \cellcolor{blue!12}{-3.73} \\  
    \cline{3-10}
    &  & \multirow{1}{*}{3-16-16} 
    & 10.00
    & 68.00
    & 74.45
    & 29.29
    & 5.97
    & 37.60
    & \cellcolor{blue!12}{-13.32} \\   
    \cline{2-10}

    & \multirow{2}{*}{\textbf{\textit{flatquant}}} 
    & \multirow{1}{*}{4-4-4} 
    & 0.00
    & 15.80
    & 16.76
    & 25.25
    & 0.00
    & 11.56
    & \cellcolor{blue!12}{-39.36} \\  
    \cline{3-10}
    &  & \multirow{1}{*}{8-8-8} 
    & 22.50
    & 82.20
    & 80.52
    & 31.31
    & 12.31
    & 45.77
    & \cellcolor{blue!12}{-5.15} \\  
    \cline{2-10}
    \hline
    \multirow{8}{*}{\rotatebox[origin=c]{90}{\textbf{\makecell{DeepSeek-R1\\-Distill-Qwen7B}}}}
    & \textbf{BF16} 
    & - 
    & \textbf{48.33} 
    & \textbf{93.20} 
    & \textbf{90.83} 
    & \textbf{48.48} 
    & \textbf{33.58} 
    & \textbf{62.89} 
    & \textbf{-} \\
    \cline{2-10}
    & \multirow{2}{*}{\textbf{\textit{awq}}} 
    & \multirow{1}{*}{4-16-16} 
    & 43.33
    & 92.80
    & 92.12
    & 47.98
    & 35.45
    & 62.34
    & \cellcolor{orange!12}{-0.55} \\  
    \cline{3-10}
    &  & \multirow{1}{*}{3-16-16} 
    & 34.17
    & 89.40
    & 90.22
    & 42.42
    & 29.10
    & 57.06
    & \cellcolor{blue!12}{-5.83} \\  
    \cline{2-10}
    
    & \multirow{1}{*}{\textbf{\textit{smoothquant}}} 
    & \multirow{1}{*}{8-8-8} 
    & 42.50
    & 91.40
    & 89.92
    & 48.99
    & 33.96
    & 61.35
    & \cellcolor{gray!12}{-1.54} \\ 
    \cline{2-10}
    
    & \multirow{2}{*}{\textbf{\textit{gptq}}} 
    & \multirow{1}{*}{4-16-16} 
    & 41.67
    & 92.40
    & 91.21
    & 47.98
    & 34.33
    & 61.52
    & \cellcolor{gray!12}{-1.37} \\  
    \cline{3-10}
    &  & \multirow{1}{*}{3-16-16} 
    & 33.33
    & 89.40
    & 90.30
    & 45.45
    & 25.00
    & 56.70
    & \cellcolor{blue!12}{-6.19} \\  
    \cline{2-10}

    & \multirow{2}{*}{\textbf{\textit{flatquant}}} 
    & \multirow{1}{*}{4-4-4} 
    & 6.67
    & 65.40
    & 79.08
    & 32.32
    & 2.24
    & 37.14
    & \cellcolor{blue!12}{-25.75} \\  
    \cline{3-10}
    &  & \multirow{1}{*}{8-8-8} 
    & 45.00
    & 95.00
    & 90.90
    & 50.00
    & 35.82
    & 63.34
    & \cellcolor{orange!12}{+0.45} \\  
    \cline{2-10}
    \hline
    
    \multirow{8}{*}{\rotatebox[origin=c]{90}{\textbf{\makecell{DeepSeek-R1\\-Distill-Qwen14B}}}}
    & \textbf{BF16} 
    & - 
    & \textbf{57.50} 
    & \textbf{95.00} 
    & \textbf{93.86} 
    & \textbf{58.08} 
    & \textbf{51.12} 
    & \textbf{71.11} 
    & \textbf{-} \\
    \cline{2-10}
    & \multirow{2}{*}{\textbf{\textit{awq}}} 
    & \multirow{1}{*}{4-16-16} 
    & 53.33
    & 95.20
    & 93.63
    & 57.58
    & 48.51
    & 69.95
    & \cellcolor{gray!20}{-1.16} \\  
    \cline{3-10}
    &  & \multirow{1}{*}{3-16-16} 
    & 44.17
    & 93.80
    & 93.71
    & 59.60
    & 42.16
    & 66.69
    & \cellcolor{blue!12}{-4.42} \\  
    \cline{2-10}
    
    & \multirow{1}{*}{\textbf{\textit{smoothquant}}} 
    & \multirow{1}{*}{8-8-8} 
    & 55.00
    & 95.40
    & 94.10
    & 56.67
    & 48.88
    & 69.90
    & \cellcolor{gray!20}{-1.21} \\ 
    \cline{2-10}
    
    & \multirow{2}{*}{\textbf{\textit{gptq}}} 
    & \multirow{1}{*}{4-16-16} 
    & 60.00
    & 95.60
    & 93.18
    & 58.08
    & 45.15
    & 70.40
    & \cellcolor{orange!20}{-0.71} \\  
    \cline{3-10}
    &  & \multirow{1}{*}{3-16-16} 
    & 47.50
    & 92.40
    & 92.42
    & 54.04
    & 43.66
    & 66.00
    & \cellcolor{blue!12}{-5.11} \\  
    \cline{2-10}

    & \multirow{2}{*}{\textbf{\textit{flatquant}}} 
    & \multirow{1}{*}{4-4-4} 
    & 4.17
    & 54.60
    & 64.97
    & 29.80
    & 7.09
    & 32.13
    & \cellcolor{blue!12}{-38.98} \\  
    \cline{3-10}
    &  & \multirow{1}{*}{8-8-8} 
    & 55.83
    & 95.20
    & 93.93
    & 59.09
    & 49.25
    & 70.66
    & \cellcolor{orange!20}{-0.45} \\  
    \cline{2-10}
    \hline
    \multirow{8}{*}{\rotatebox[origin=c]{90}{\textbf{\makecell{QwQ-32B}}}}
    & \textbf{BF16} 
    & - 
    & \textbf{71.67} 
    & \textbf{97.00} 
    & \textbf{95.30} 
    & \textbf{63.64} 
    & \textbf{61.19} 
    & \textbf{81.29} 
    & \textbf{-} \\
    \cline{2-10}
    & \multirow{2}{*}{\textbf{\textit{awq}}} 
    & \multirow{1}{*}{4-16-16} 
    & 77.50
    & 96.80
    & 96.13
    & 62.12
    & 60.07
    & 78.53
    & \cellcolor{gray!12}{-2.76} \\  
    \cline{3-10}
    &  & \multirow{1}{*}{3-16-16} 
    & 67.50
    & 96.00
    & 94.84
    & 59.09
    & 48.88
    & 73.56
    & \cellcolor{blue!12}{-7.73} \\  
    \cline{2-10}
    
    & \multirow{1}{*}{\textbf{\textit{smoothquant}}} 
    & \multirow{1}{*}{8-8-8} 
    & 78.33
    & 97.60
    & 95.98
    & 67.17
    & 60.45
    & 79.91
    & \cellcolor{gray!12}{-1.38} \\ 
    \cline{2-10}
    
    & \multirow{2}{*}{\textbf{\textit{gptq}}} 
    & \multirow{1}{*}{4-16-16} 
    & 73.33
    & 97.20
    & 95.60
    & 61.11
    & 56.34
    & 76.72
    & \cellcolor{blue!12}{-4.57} \\  
    \cline{3-10}
    &  & \multirow{1}{*}{3-16-16} 
    & 60.00
    & 96.20
    & 95.30
    & 54.04
    & 48.51
    & 70.81
    & \cellcolor{blue!12}{-10.48} \\  
    \cline{2-10}

    & \multirow{2}{*}{\textbf{\textit{flatquant}}} 
    & \multirow{1}{*}{4-4-4} 
    & 40.83
    & 94.00
    & 94.62
    & 56.06
    & 38.43
    & 64.79
    & \cellcolor{blue!12}{-16.50} \\  
    \cline{3-10}
    &  & \multirow{1}{*}{8-8-8} 
    & 75.00
    & 97.40
    & 95.38
    & 62.63
    & 57.49
    & 77.57
    & \cellcolor{blue!12}{-3.7} \\  
    \cline{2-10}
    \hline\hline
    \end{tabular}
}
\caption{
Performance of quantized DeepSeek-R1-Distill-Qwen models ((1.5B, 7B, 14B)) and QwQ-32B models on the Ascend NPU across various reasoning benchmarks using default hyperparameters. 
The \textcolor[HTML]{F0C869}{orange}, \textcolor[HTML]{A3A19D}{gray} and \textcolor[HTML]{9C9DDB}{blue} cells stand for the lossless (\textcolor[HTML]{F0C869}{$\leq$1\%}), the fair (\textcolor[HTML]{A3A19D}{1\%-3\%}) and the risky (\textcolor[HTML]{9C9DDB}{$\geq$3\%}) respectively.
}
\label{tab:reason}
\end{table*}
Regarding weight-only quantization, we confirm that 4-bit configurations (e.g., AWQ and GPTQ with w4g128) maintain relatively acceptable performance, with accuracy drops generally within 4.5\% of the full-precision (BF16) baseline. However, consistent with observations in \citep{liu2025quantization}, reducing the precision to 3-bit (w3g128) leads to a prohibitive performance collapse, particularly in smaller models. 

For weight-activation quantization, our evaluation of the SmoothQuant algorithm (W8A8KV8) similarly demonstrates "fair" stability, also staying within a 4.5\% margin of the baseline. Nevertheless, a systematic performance gap persists: across all tested algorithms (AWQ, GPTQ, and SmoothQuant), the results obtained on the Ascend NPU are consistently inferior to the GPU-based metrics reported in \citep{liu2025quantization}. This suggests that while the algorithmic trends are cross-platform, the specific hardware execution on the NPU introduces additional numerical overhead that hinders the models from reaching their full theoretical potential under low-bit precision.

Our evaluation on the Ascend NPU largely corroborates the findings in \citep{liu2025quantization}. AWQ generally performs on par with GPTQ across most model scales. A notable exception is the DeepSeek-R1-Distill-Qwen-1.5B model, where AWQ exhibits a higher accuracy drop than GPTQ. However, it should be noted that at the 1.5B scale, both algorithms incur significant degradation, whereas severe performance collapse is reserved for 3-bit configurations.

Given its overall robustness in larger models and superior implementation efficiency (avoiding the iterative parameter updates required by GPTQ), we align with the recommendation of AWQ as the default weight-only quantization algorithm for reasoning models on the Ascend NPU.
\begin{table*}[!t]
\centering 
\resizebox{1.00\linewidth}{!}{
    \begin{tabular}{c|l|l|rrrrrr|r|r}  
    \hline\hline
    \textbf{Model} 
    & \textbf{Methods} 
    & \textbf{\makecell{W-A-KV\\\# Bits}} 
    & \textbf{\makecell{ARC-C}}  
    & \textbf{\makecell{ARC-E}} 
    & \textbf{HellaSwag} 
    & \textbf{\makecell{LAMBADA}} 
    & \textbf{\makecell{PIQA}} 
    & \textbf{\makecell{Winogrande}} 
    & \textbf{Avg.} 
    & \textbf{Drop}$\downarrow$ \\ 
    \hline
    \multirow{8}{*}{\rotatebox[origin=c]{90}{\textbf{\makecell{LLaMA-3-8B}}}}
    & \textbf{BF16-X2000} 
    & -
    & \textbf{53.50} 
    & \textbf{77.57} 
    & \textbf{79.12} 
    & \textbf{75.51} 
    & \textbf{80.74}
    & \textbf{72.93} 
    & \textbf{73.23} 
    & \textbf{-} \\
    \cline{2-11}
    
    & \multirow{2}{*}{\textbf{\textit{flatquant-X2000}}} 
    & \multirow{1}{*}{8-8-8} 
    & 53.33
    & 77.53
    & 79.12
    & 75.68
    & 80.63
    & 73.40
    & 73.28
    & \cellcolor{orange!20}{+0.05   } \\  
    \cline{3-11}
    &  & \multirow{1}{*}{4-4-4}
    & 50.00
    & 75.80
    & 76.80
    & 72.91
    & 79.16
    & 72.69
    & 71.23
    & \cellcolor{gray!20}{-2.00} \\  
    \cline{2-11}
    & \textbf{BF16-NPU} 
    & - 
    & \textbf{54.18}
    & \textbf{77.86}
    & \textbf{79.20}
    & \textbf{75.55}
    & \textbf{80.36}
    & \textbf{72.69} 
    & \textbf{73.31}
    & \textbf{-} \\
    \cline{2-11}
    
    & \multirow{4}{*}{\textbf{\textit{flatquant-NPU}}} 
    & \multirow{1}{*}{8-8-8}
    & 53.24
    & 77.95
    & 79.08
    & 75.37
    & 80.36
    & 72.14
    & 73.02
    & \cellcolor{orange!20}{-0.29} \\  
    \cline{3-11}
    &  & \multirow{1}{*}{4-4-4$^{*}$} 
    & 25.17
    & 25.04
    & 25.19
    & 0.00
    & 51.25
    & 48.62
    & 29.21
    & \cellcolor{blue!20}{-44.02} \\  
    \cline{3-11}
    &  & \multirow{1}{*}{4-4-4$^{\dagger}$} 
    & 45.14
    & 70.08
    & 73.13
    & 59.31
    & 75.19
    & 65.75
    & 64.77
    & \cellcolor{blue!20}{-8.46} \\  
    \cline{3-11}
    &  & \multirow{1}{*}{4-4-4$^{\ddagger}$} 
    & 48.04
    & 71.46
    & 74.83
    & 68.12
    & 74.92
    & 65.75
    & 67.19
    &  \cellcolor{blue!20}{-6.04} \\  
    \cline{2-11}
    \hline\hline
    \end{tabular}
}
\caption{
The overall assessment of quantized LLaMA-3-8B model with X2000 and NPU on various QA benchmarks. 
The \textcolor[HTML]{F0C869}{orange}, \textcolor[HTML]{A3A19D}{gray} and \textcolor[HTML]{9C9DDB}{blue} cells stand for the lossless (\textcolor[HTML]{F0C869}{$\leq$1\%}), the fair (\textcolor[HTML]{A3A19D}{1\%-3\%}) and the risky (\textcolor[HTML]{9C9DDB}{$\geq$3\%}) respectively.
Quantization variants are defined as follows:
$^{*}$~denotes the original FlatQuant configuration with default calibration hyperparameters;
$^{\dagger}$~denotes an adjusted set of calibration hyperparameters that achieves better performance;
$^{\ddagger}$~denotes the best-performing configuration obtained via multi-layer hyperparameter tuning, derived from extensive empirical exploration and a layer-wise parameter adjustment strategy.
}

\label{tab:QA}
\end{table*}

Our experiments reveal a pronounced performance gap between FlatQuant-W8A8KV8 and FlatQuant-W4A4KV4 configurations on the Ascend NPU when utilizing the original hyperparameters from the official codebase \citep{flatquant_code, qhs_code}:
\begin{itemize}
    \item \textbf{8-bit Quantization (W8A8KV8)}: This configuration maintains stable PPL and downstream task performance on the Ascend NPU in Table \ref{tab:reason}, closely aligning with GPU-based results and reported baselines \citep{liu2025quantization}. These results hold across both short-form question answering and long-context reasoning benchmarks, suggesting that W8A8KV8 quantization effectively preserves model integrity across diverse task types and sequence lengths.
    \item \textbf{4-bit Quantization}: We observe divergent behaviors depending on the task category. On QA benchmarks, while initial results indicated significant degradation, targeted hyperparameter tuning (notably per-layer learning-rate adjustment and epoch selection) recovers performance to near the original FlatQuant baseline, with a small remaining gap (Table \ref{tab:QA}). 
    On reasoning benchmarks with long contexts, DeepSeek-R1-Distill-Qwen models (1.5B, 7B, and 14B) suffer severe quality loss; even after extensive tuning, \textit{logic collapse} persists, manifested by extreme PPL spikes and catastrophic generation failures (e.g., missing EOS tokens). During calibration, sustained MSE oscillations and outlier layers further indicate that 4-bit weight-activation quantization can be unstable for these reasoning-intensive workloads on Ascend NPUs. In contrast, QwQ-32B exhibits a smaller and more tolerable accuracy drop in Table \ref{tab:reason}, but it still underperforms the results reported in QHS\citep{liu2025quantization}.
\end{itemize}

It is noteworthy that our calibration and evaluation are conducted using fake quantization, meaning the underlying computations remain in higher precision. Therefore, the observed performance gap suggests that the numerical sensitivity of 4-bit quantization schemes is further amplified by platform-specific operator implementations on the Ascend NPU. 
Overall, Ascend NPU shows reliable numerical behavior at 8-bit precision, while 4-bit quantization across weights, activations, and KV caches remains challenging, especially for long-generation reasoning tasks.

\subsection{Comparison of FlatQuant-W4A4KV4 Between NPU and X2000}

\begin{figure}[t]
    \centering
    \begin{subfigure}[b]{0.42\textwidth}
        \centering
        \includegraphics[width=\textwidth]{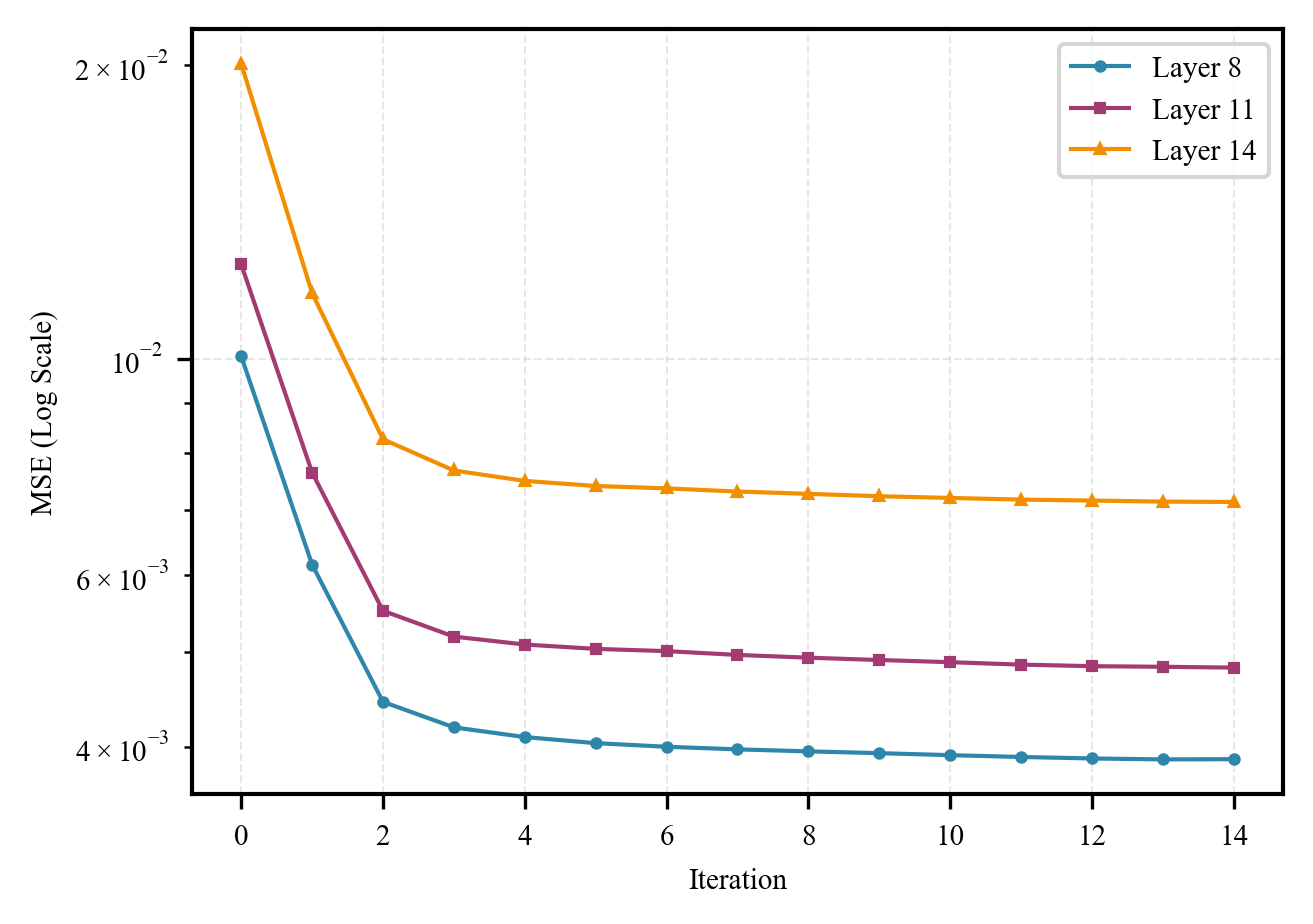}
        \caption{Llama3-8B (X2000)}
        \label{fig:X2000_llama3}
    \end{subfigure}
    \hfill
    \begin{subfigure}[b]{0.42\textwidth}
        \centering
        \includegraphics[width=\textwidth]{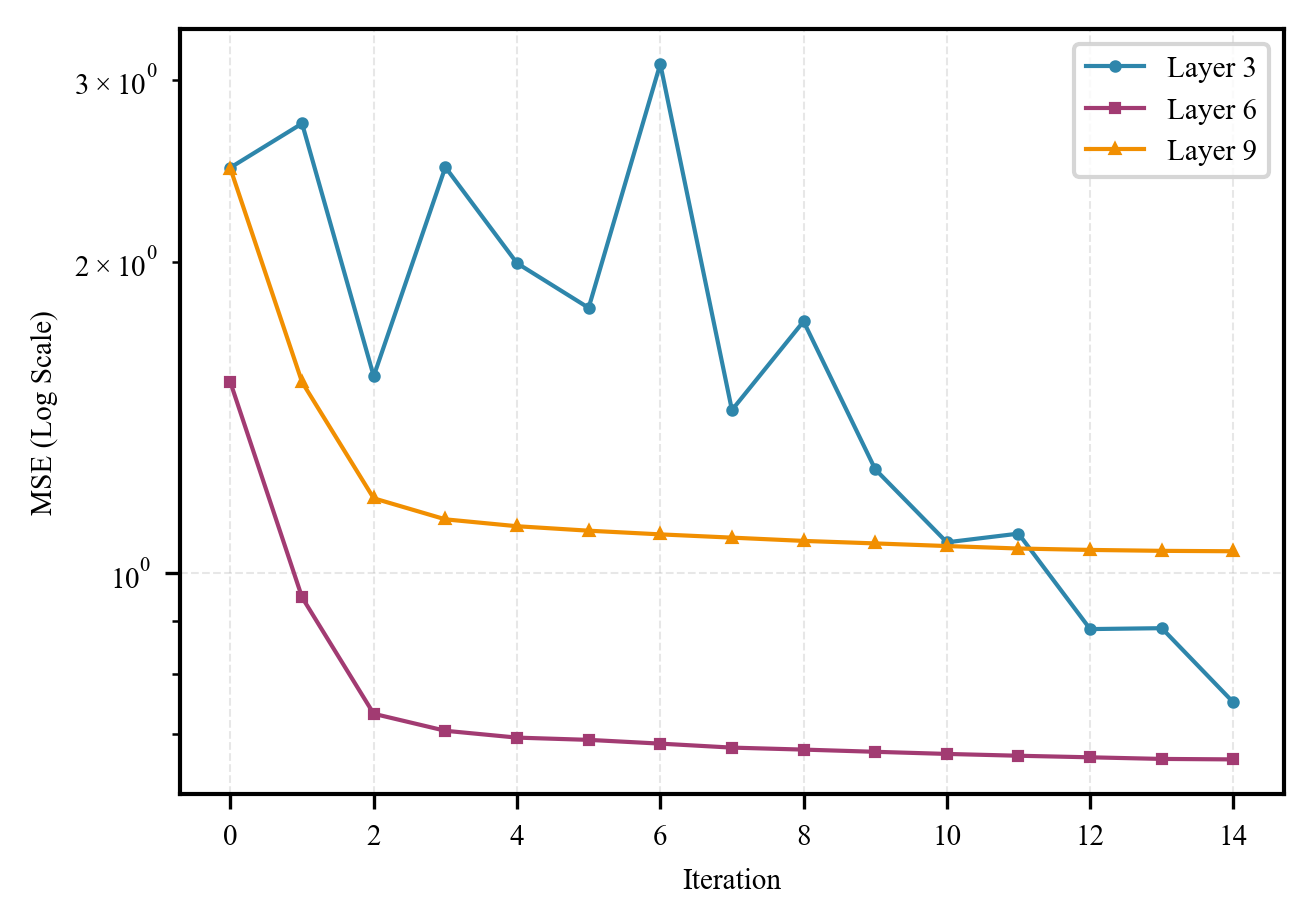}
        \caption{DeepSeek-R1-Distill-Qwen-7B (X2000)}
        \label{fig:X2000_qwen}
    \end{subfigure}
    
    \vspace{1em} 
    
    \begin{subfigure}[b]{0.42\textwidth}
        \centering
        \includegraphics[width=\textwidth]{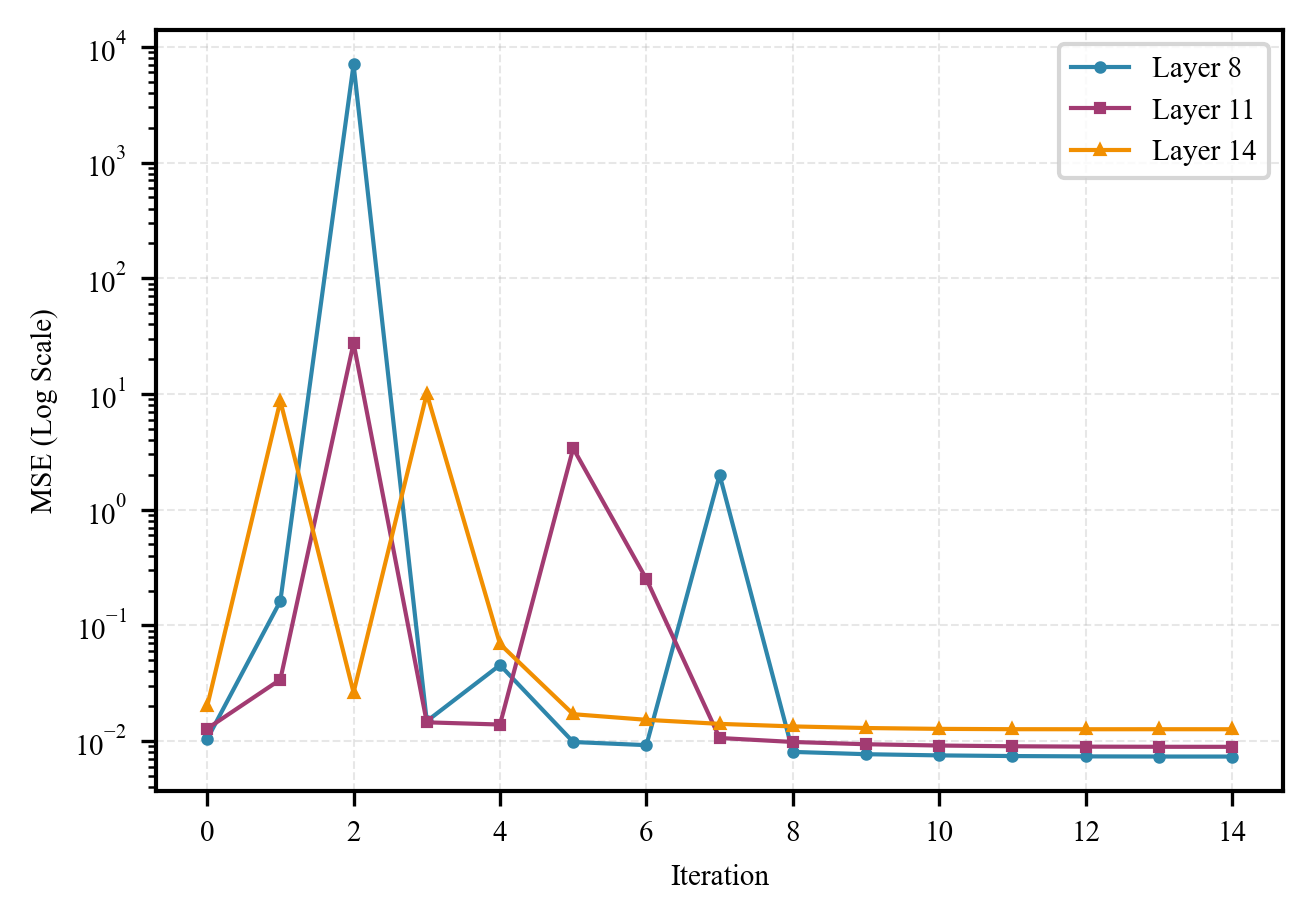}
        \caption{Llama3-8B (NPU)}
        \label{fig:npu_llama3}
    \end{subfigure}
    \hfill
    \begin{subfigure}[b]{0.42\textwidth}
        \centering
        \includegraphics[width=\textwidth]{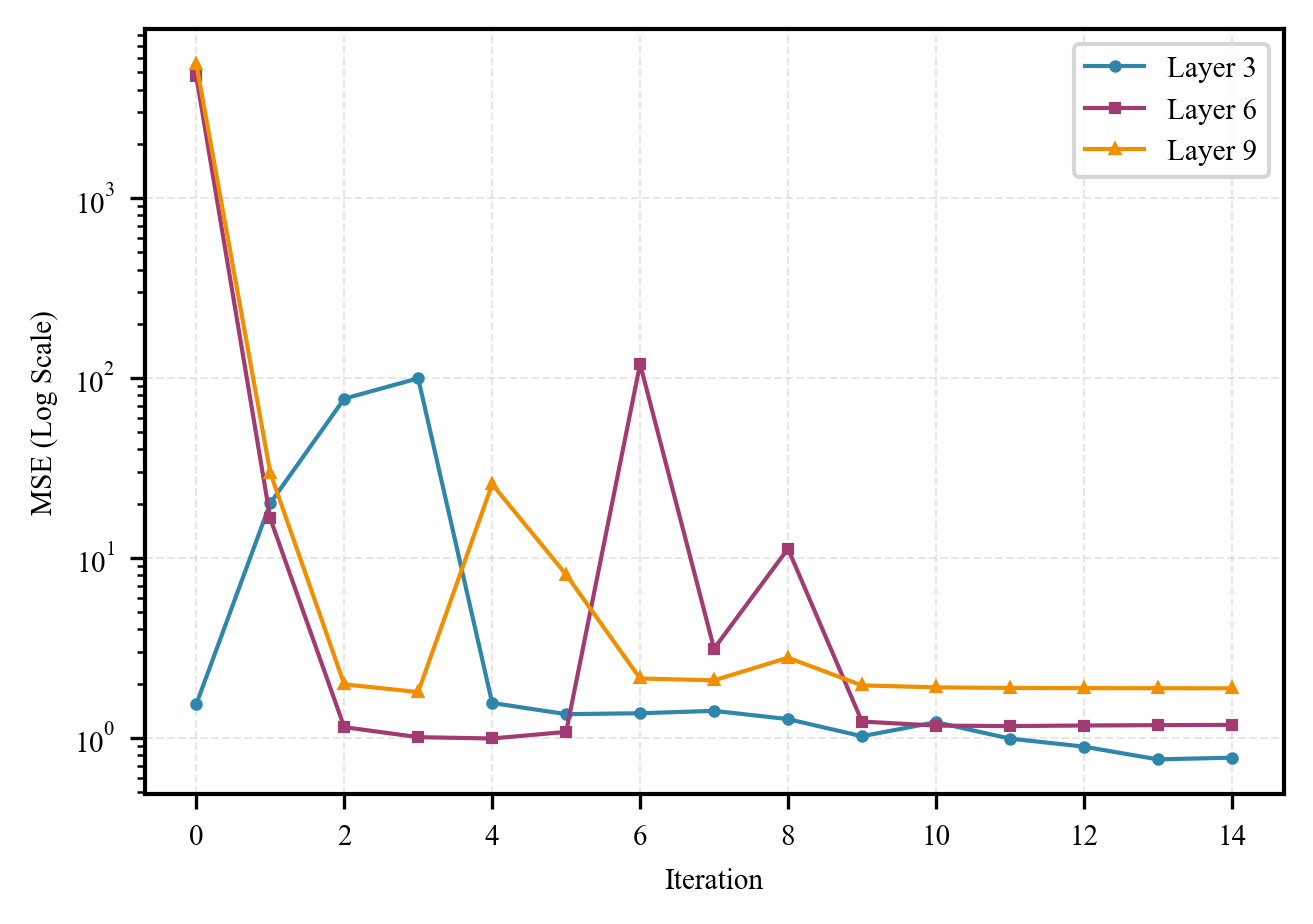}
        \caption{DeepSeek-R1-Distill-Qwen-7B (NPU)}
        \label{fig:npu_qwen}
    \end{subfigure}
    
        \caption{Comparison of layer-wise MSE loss for FlatQuant (W4A4KV4). From left to right: Llama3-8B and DeepSeek-R1-Distill-Qwen-7B on X2000, followed by the same models on NPU. The results illustrate a significant increase in quantization error when transitioning from X2000 to NPU hardware.}
    \label{fig:mse_loss_comparison}
\end{figure}
    

Given the performance gap between NPU and X2000 implementations of W4A4KV4 quantization, we reproduced the 4-bit FlatQuant experiments on an X2000 as comparison and conducted a granular layer-wise analysis of the Mean Squared Error (MSE) loss to pinpoint the discrepancies.
All experiments strictly follow the original FlatQuant implementation and the recommended hyperparameters used in the official code \citep{flatquant_code, qhs_code}. 

\textbf{4-bit FlatQuant on X2000.} Under these settings, both Llama-3-8B and DeepSeek-R1-Distill-Qwen-7B exhibit stable behavior throughout the calibration process.
In particular, the layer-wise MSE decreases smoothly without large oscillations or outlier layers (Figure \ref{fig:mse_loss_comparison}).
After calibration, the quantized models achieve reasonable PPL values that are consistent with reported baselines.
Importantly, no layer-wise tuning or heterogeneous hyperparameter adjustment is required.
A single global configuration is sufficient to obtain stable 4-bit quantization results on X2000.

\textbf{4-bit FlatQuant on NPU.} In contrast, deploying the same 4-bit FlatQuant algorithm with identical hyperparameters on the Ascend NPU yields markedly different results, characterized by significant numerical instability.
First, during the calibration phase, we observe erratic optimization behavior: the layer-wise MSE fails to converge smoothly, exhibiting large oscillations across training steps (Figure \ref{fig:mse_loss_comparison}). 
Second, this calibration instability propagates to the final model, precipitating a catastrophic degradation in inference accuracy. The instability is particularly detrimental during long-context reasoning tasks. For instance, in models such as DeepSeek-R1-Distill-Qwen-7B, 4-bit quantization on the NPU frequently induces abnormal generation behaviors, characterized by the failure to emit end-of-sequence (EOS) tokens and the onset of repetitive, infinite generation loops.

Cross-platform comparisons suggest that the severe degradation observed on the Ascend NPU is not solely caused by the high-level FlatQuant algorithm itself. Instead, the instability likely relates to platform-specific numerical behaviors during simulated low-precision computation. While the simulated 4-bit quantization remains stable when executed on X2000, the identical configuration exposes significant numerical fragility on the NPU. This suggests that the discrepancy arises from how different hardware backends handle the stochasticity or precision loss inherent in fake quantization, rather than from the high-level algorithmic logic.

\subsection{Mitigating Calibration Fragility of W4A4KV4 Rotations on Ascend NPUs}

\textbf{ Non-Linear Dominance of Layer-wise MSE.}
A closer inspection of the calibration traces suggests that failure is not caused by a gradual accumulation of small errors across all layers. Instead, the final PPL is often dominated by one or a few layers whose quantization error becomes abnormally large. Empirically, even when most layers achieve low MSE after calibration, a single layer with an extreme MSE value can still cause catastrophic degradation in both PPL and downstream task performance.

\textbf{Independent layer-wise calibration objective.}
This behavior can be explained by the layer-wise objective used in FlatQuant calibration.
During calibration, quantization parameters are optimized independently for each layer by minimizing:
\begin{equation*}
\min_{\theta^{(l)}} \frac{1}{B} \sum_{j=1}^{B} \mathcal{L} \left( \text{Layer}_{fp}^{(l)}(\mathbf{x}_j), \text{Layer}_{quant}^{(l)}(\mathbf{x}_j; \theta^{(l)}) \right)
\end{equation*}
where $\mathbf{x}_j$ denotes the input activations for the $j$-th batch and $\theta^{(l)}$ denotes learnable quantization parameters (e.g., diagonal scaling factors or clipping thresholds). We use the MSE reconstruction loss:
\begin{equation*}
\mathcal{L} = \| \mathbf{Y}_{fp}^{(l)} - \hat{\mathbf{Y}}_{quant}^{(l)} \|_2^2.
\end{equation*}

The parameters are updated iteratively using AdamW over $E$ epochs.
Since each layer is optimized in isolation, the objective only measures the local approximation error of the current layer; errors from earlier layers are not exposed to the optimizer, and later layers cannot reliably compensate for severe numerical errors in earlier layers. Consequently, once a sensitive layer converges to a poor solution, the induced large MSE tends to persist and can dominate the final model quality.

\textbf{Fine-grained Layer-wise Calibration Strategy.} Guided by these insights, we further analyzed layer-wise MSE and identified a small set of sensitive layers in Llama-3-8B. We then applied a refined calibration strategy with layer-specific learning rates and extended optimization epochs. As summarized in Table \ref{tab:QA}, this targeted refinement narrows the performance gap between Ascend NPU and the X2000 baseline, although a small discrepancy remains, suggesting that additional low-level optimizations may be required for full parity. 



\subsection{Real-World Deployment and Acceleration on Ascend NPU}
 \textbf{Simulated Fake Quantization Protocol.} For the majority of our comparative evaluations, we adopt a unified fake-quantization strategy. This approach simulates the rounding effects and clipping errors introduced by low-precision hardware while maintaining tensors in floating-point format for consistent back-propagation and metric calculation.

Given a high-precision input $x$, the simulated quantized value $\hat{x}$ is formally defined as:
\begin{equation*}
\hat{x} = \text{Dequant}(\text{Quant}(x; S, Z)) = \left( \text{clamp}\left( \left\lfloor \frac{x}{S} \right\rceil + Z, q_{min}, q_{max} \right) - Z \right) \times S
\end{equation*}
where $S$ denotes the quantization scaling factor and $Z$ represents the integer zero-point. The operator $\lfloor \cdot \rceil$ indicates rounding to the nearest integer, and $\text{clamp}(\cdot)$ confines values within the representable range $[q_{min}, q_{max}]$.

In the context of FlatQuant, this differentiable simulation enables the AdamW optimizer to minimize reconstruction error by iteratively adjusting $S$ and clipping thresholds. Specifically for activations, we model the learned clipping mechanism as:
\begin{equation*}
x_{\text{clip}} = \mathrm{clip}(x, -\alpha, \alpha)
\end{equation*}
While this proxy effectively captures theoretical precision loss, it does not reflect actual hardware latency, serving primarily as an algorithmic verification tool.

To strictly validate the hardware capabilities of the Ascend NPU, we moved beyond simulation for the SmoothQuant (W8A8KV8) configuration, deploying a fully functional real-quantization inference pipeline.

Unlike the simulated baselines, this deployment integrates the native dynamic quantization operators from \texttt{torch\_npu} and leverages high-performance INT8$\times$INT8 matrix multiplication kernels \citep{torchnpu_npu_quant_matmul}. A key kernel is fused kernel of add, softmax and drop, whose calculation logic is:

$
\begin{aligned}
\textbf{Inputs:}& \\
& \mathbf{X} \in \mathbb{R}^{m \times n} \quad \text{(Primary input tensor)} \\
& \mathbf{M} \in \mathbb{R}^{m \times n} \quad \text{(Tensor to be added)} \\
\text{Parameters:}& \\
& p \in [0, 1) \quad \text{(Dropout probability)} \\
& \text{dim} \quad \text{(Dimension for Softmax, typically } -1 \text{)} \\
\text{Process:}& \\
& \mathbf{S} = \mathbf{X} + \mathbf{M} \\
& \mathbf{A}_i = \frac{e^{S_i}}{\sum_{j}^{\text{dim}} e^{S_j}} \quad \text{(Softmax along specified dimension)} \\
& \mathbf{R} \sim \text{Bernoulli}(1-p) \quad \text{(Random mask, where } R_i \in \{0, 1\}) \\
& \mathbf{Y}_i = \mathbf{R}_i \cdot \frac{\mathbf{A}_i}{1-p} \quad \text{(Apply dropout mask and scaling)} \\
\textbf{Output:}& \\
& \mathbf{Y} \in \mathbb{R}^{m \times n} \quad \text{(Fused result tensor)}
\end{aligned}
$

By registering these kernels within the \texttt{vllm-ascend} backend, we successfully instantiated a real-quantization inference pipeline, enabling the actual execution of the quantized model on hardware.
\begin{table}[htbp]
\centering
\caption{Inference Performance Comparison of DeepSeek-R1-Distill-Qwen7B: Pseudo vs. Real Quantization}
\label{tab:speedup}
\renewcommand{\arraystretch}{1.3} 

\resizebox{\textwidth}{!}{%
\begin{tabular}{l c c cc cc cc cc cc c}
\toprule
\multirow{2}{*}{\textbf{Method}} & \multirow{2}{*}{\textbf{PPL}} & \multirow{2}{*}{\textbf{Seed}} & \multicolumn{2}{c}{\textbf{AIME-120}} & \multicolumn{2}{c}{\textbf{MATH-500}} & \multicolumn{2}{c}{\textbf{GSM8K}} & \multicolumn{2}{c}{\textbf{GPQA-Diamond}} & \multicolumn{2}{c}{\textbf{LiveCodeBench}} & \textbf{Average} \\
\cmidrule(lr){4-5} \cmidrule(lr){6-7} \cmidrule(lr){8-9} \cmidrule(lr){10-11} \cmidrule(lr){12-13} \cmidrule(lr){14-14}
 & & & Acc & Time (h:m:s) & Acc & Time (h:m:s) & Acc & Time (h:m:s) & Acc & Time (h:m:s) & Acc & Time (h:m:s) & Acc \\
\midrule

\textbf{Original Model} & 
26.01 & 42 & 
42.5 & 1:10:05 & 
93.60 & 56:55 & 
91.05 & 47:00 & 
52.53 & 32:26 & 
36.19 & 53:26 & 
63.17 \\
\midrule

\textbf{SmoothQuant-W8A8KV8} & \multirow{2}{*}{26.09} & \multirow{2}{*}{42} & \multirow{2}{*}{43.33} & \multirow{2}{*}{4:08:49} & \multirow{2}{*}{91.60} & \multirow{2}{*}{3:03:24} & \multirow{2}{*}{90.37} & \multirow{2}{*}{1:43:56} & \multirow{2}{*}{48.48} & \multirow{2}{*}{01:37:09} & \multirow{2}{*}{35.82} & \multirow{2}{*}{3:05:11} & \multirow{2}{*}{61.92} \\
{\footnotesize (Pseudo-Quant / Original Code)} & & & & & & & & & & & & & \\
\midrule

\begin{tabular}[c]{@{}l@{}}
\textbf{SmoothQuant-W8A8KV8} \\
{\footnotesize (Optimized Kernel)}
\end{tabular} & 
26.09 & 42 & 
37.5 & 1:47:02 & 
94.20 & 1:17:53 & 
89.46 & 1:23:17 & 
51.01 & 1:03:09 & 
35.45 & 1:18:25 & 
61.52 \\

\bottomrule
\end{tabular}
}
\vspace{0.1cm}
\footnotesize{
\textbf{Note}: 
PPL is evaluated on the WikiText2 dataset. 
Time represents inference latency, formatted as h:m:s (hours:minutes:seconds). Both quantized models are calibrated using the Numina-Math-1.5 \citep{li2024numinamath}.
}
\end{table}

\textbf{Performance and Hardware Efficacy.} While real quantization theoretically offers speedups, our practical deployment on the Ascend NPU reveals a latency trade-off. As shown in Table \ref{tab:speedup}, the integration of the optimized INT8 matrix multiplication kernel significantly reduces latency compared to the pseudo-quantization baseline (e.g., in AIME-120). However, the end-to-end inference speed of the quantized model remains slower than the original full-precision baseline.
This performance gap is likely attributed to the overhead of dynamic quantization operators. Specifically, operations following the KV-cache currently necessitate on-the-fly quantization, which introduces additional latency. While a comprehensive profiling is reserved for future work, these results suggest that although the NPU demonstrates strong potential for accelerating low-bit matrix multiplications, the current overhead from discrete quantization ops partially offsets these computational gains.

\section{Discussion}

\paragraph{Platform sensitivity of low-bit quantization.}
Across both QA and reasoning benchmarks, we observe that the same PTQ algorithm and nominal bit-width can lead to noticeably different outcomes on Ascend NPUs compared with GPU-reported results.
This gap is particularly visible under more aggressive settings (e.g., 4-bit W-A-KV quantization), suggesting that \textit{algorithmic correctness} alone is insufficient to guarantee stable behavior when the underlying operator implementations and numerical details change.
From a deployment perspective, this implies that conclusions drawn purely from GPU experiments should be re-validated on the target NPU backend.

\paragraph{Why does 4-bit W-A-KV quantization fail for long-context reasoning on Ascend?}
Our experiments show that FlatQuant-W4A4KV4 can be made competitive on QA tasks after targeted hyperparameter tuning, yet remains fragile for long-context reasoning on distilled models, where ``logic collapse'' (e.g., abnormal generation behavior) is frequently observed.
A layer-wise analysis suggests that failures are dominated by a small number of sensitive layers whose quantization error becomes abnormally large, which can overwhelm downstream layers and lead to catastrophic quality degradation.
This observation motivates future work on (i) more robust calibration objectives that expose cross-layer effects, and (ii) NPU-aware quantization recipes that explicitly account for backend numerical characteristics.

\section{Conclusion}

We presented an empirical evaluation of representative post-training quantization methods on Huawei Ascend NPUs, covering both QA-focused and reasoning-focused benchmarks.
Our results show that 4-bit weight-only quantization can be viable for sufficiently large models, whereas pushing to 3-bit often causes severe accuracy collapse.
At 8-bit precision, SmoothQuant and FlatQuant-W8A8KV8 remain numerically stable on Ascend across a broad set of tasks.
In contrast, aggressive 4-bit W-A-KV quantization (FlatQuant-W4A4KV4) exhibits strong platform sensitivity and can become unstable for long-context reasoning workloads.
Beyond accuracy, we also implemented a partial real-quantized INT8 inference path to study practical acceleration on Ascend.
While optimized INT8$\times$INT8 kernels can reduce latency for certain components, dynamic quantization overheads currently limit end-to-end speedups, indicating that further operator coverage and kernel/runtime co-optimization are necessary.
Overall, our findings provide evidence that NPU backends introduce non-trivial numerical and systems factors that can materially affect low-bit quantization outcomes, and we hope they can serve as a reference for future NPU-oriented quantization research and deployment.
    
\bibliographystyle{plainnat}  
\bibliography{references}  


\end{document}